%% file: main.tex
\pdfoutput=1

\documentclass[11pt]{article}

\usepackage[final]{acl}

\usepackage{times}
\usepackage{latexsym}

\usepackage[T1]{fontenc}

\usepackage[utf8]{inputenc}

\usepackage{microtype}

\usepackage{inconsolata}

\usepackage{graphicx}

\usepackage{xcolor}
\usepackage{booktabs}  
\usepackage{graphicx}  
\usepackage{array}     
\usepackage{caption}   
\usepackage{multirow}
\usepackage{amsfonts}
\usepackage{amsmath}
\usepackage{amsthm}
\usepackage{amssymb}
\usepackage{tabularx} 

\definecolor{maic-color}{RGB}{118,83,179} 

\newtheorem{definition}{Definition}

\newcommand{\vpara}[1]{\vspace{0.05in}\noindent \textbf{#1 }}

\title{Handling Students Dropouts in an LLM-driven \\ Interactive Online Course Using Language Models}

\author{Yuanchun Wang$^{\spadesuit}$\footnote{}, Yiyang Fu$^{\spadesuit}$\footnotemark[1], Jifan Yu$^\clubsuit$, Daniel Zhang-Li$^\diamondsuit$, Zheyuan Zhang$^\heartsuit$, \\ 
\textbf{Joy Lim Jia Yin$^\diamondsuit$, Yucheng Wang$^\diamondsuit$, Peng Zhou$^\diamondsuit$, Jing Zhang$^{\spadesuit}$, Huiqin Liu$^\clubsuit$} \\
$^{\spadesuit}$School of Information, Renmin University of China\ \ \ \ \ \ 
\\$^\clubsuit$Institute of Education, Tsinghua University\ \ \ \ \ \
\\$^\diamondsuit$Department of Computer Science and Technology, Tsinghua University\ \ \ \ \ \
\\$^\heartsuit$Language Technology Institute, Carnegie Mellon University\ \ \ \ \ \
}

\begin{document}
\maketitle

\renewcommand{\thefootnote}{\fnsymbol{footnote}}
\footnotetext[1]{Both authors contributed equally to this research.}
\renewcommand*{\thefootnote}{\arabic{footnote}}

\input{paragraphs/0_abstract}
\input{paragraphs/1_introduction}
\input{paragraphs/2_related_works}
\input{paragraphs/3_dropouts_analysis}
\input{paragraphs/4_dropous_prediction}
\input{paragraphs/5_dropous_intervention}
\input{paragraphs/6_conclusion}

\input{paragraphs/7_limitations_and_ecs}

\bibliography{custom}

\input{paragraphs/8_appendix}
\end{document}

%% file: paragraphs/0_abstract.tex

\begin{abstract}
Interactive online learning environments, represented by Massive AI-empowered Courses~(MAIC), leverage LLM-driven multi-agent systems to transform passive MOOCs into dynamic, text-based platforms, enhancing interactivity through LLMs. 
This paper conducts an empirical study on a specific MAIC course to explore three research questions about dropouts in these interactive online courses: (1) What factors might lead to dropouts? (2) Can we predict dropouts? (3) Can we reduce dropouts?  
We analyze interaction logs to define dropouts and identify contributing factors. 
Our findings reveal strong links between dropout behaviors and textual interaction patterns. 
We then propose a course-progress-adaptive dropout prediction framework~(CPADP) to predict dropouts with at most 95.4\% accuracy. 
Based on this, we design a personalized email recall agent to re-engage at-risk students.  

Applied in the deployed MAIC system with over 3,000 students, the feasibility and effectiveness of our approach have been validated on students with diverse backgrounds.  
\end{abstract}  

%% file: paragraphs/1_introduction.tex
\section{Introduction}

Massive Open Online Courses (MOOCs) have become a widely adopted form of online education~\cite{anders2015theories}.
In a typical MOOC setting, learners primarily acquire knowledge by watching prerecorded instructional videos online~\cite{baturay2015overview}. 
\begin{figure}[t]
    \centering
    \includegraphics[width=\linewidth]{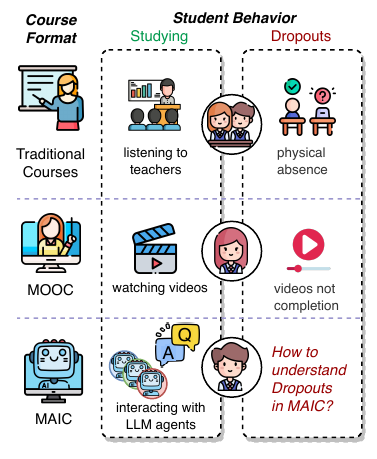}
    \caption{Dropouts in traditional courses (top), MOOC (middle), and MAIC (bottom).}
    \label{fig:intro}
\end{figure}
However, MOOCs worldwide face a critical and persistent challenge: extremely high dropout rates~\cite{feng2019understanding}. 
Dropouts in MOOCs refer to learners not completing the full sequence of instructional videos or course activities, similar to physical absence in traditional classrooms, as shown in Figure~\ref{fig:intro}. 
Despite various efforts to reduce dropout rates, the passive, video-based learning approach of MOOCs makes it difficult to maintain learners' attention and offer interactions as engaging as those in traditional classrooms~\cite{hew2014students}.

Interactive online learning environments~\cite{alsafari2024towards,tan2025elf}, represented by Massive AI-empowered Courses (MAIC)~\cite{yu2024moocmaicreshapingonline}, use LLM-driven multi-agent systems to transform passive MOOCs into dynamic, interactive learning experiences.
Unlike traditional MOOCs with passive video consumption, MAIC enables interactive, text-based learning through conversations with LLM agents guided by instructional slides~\cite{zhang2024awaking}.
This new paradigm of online interaction compels learners to actively engage with course content and maintain continuous interaction throughout the learning process\cite{zhang2024simulating}.

However, it remains uncertain whether previous research conclusions on dropouts in MOOCs can be directly transferred to MAIC.
How to define dropouts and the following key research questions need to be answered in MAIC:

\hspace{1em} \vpara{\textcolor{maic-color}{RQ~1:}} \textit{What might lead to dropouts?}

\hspace{1em} \vpara{\textcolor{maic-color}{RQ~2:}} \textit{Can we predict dropouts?}

\hspace{1em} \vpara{\textcolor{maic-color}{RQ~3:}} \textit{Can we reduce dropouts?}

To address these questions, we conducted an empirical study on a specific MAIC course, \textit{Towards Artificial General Intelligence}, which focuses on AI development and language models across six well-structured chapters.  
We collect detailed log data from course interactions and perform an exploratory analysis to define and identify dropouts and their contributing factors.  
Our preliminary findings reveals that dropouts in MAIC are strongly associated with textual interaction patterns between learners and AI agents.  
Building on this insight, we formally define the dropout prediction task in MAIC and propose a \textbf{C}ourse-\textbf{P}rogress-\textbf{A}daptive \textbf{D}ropouts \textbf{P}rediction~(CPADP) framework, to predict dropouts.  
Based on the analysis and prediction, we further design a personalized email recall agent that generates unique, tailored emails for each student based on their individual interaction records, aiming to re-engage students at risk of dropping out and rekindle their interest in the course to encourage continued learning.

The prediction and intervention system is implemented in the real-world applied MAIC system.
Tests on the dataset constructed on the MAIC-TAGI course shows that the CPADP framework has a theoretical accuracy of at most 95.4\%.
The feasibility of the personalized Email intervention approach are also validated in this course in a new semester.

%% file: paragraphs/2_related_works.tex
\section{Related Works}
\subsection{Dropouts in Traditional Online Courses}
Dropouts in traditional online courses, represented by MOOCs, have been a long-standing challenge, with reports showing the dropout rates in MOOCs worldwide reached over 95\%~\cite{feng2019understanding}. 

Understanding why learners drop out of MOOCs is critical for addressing the issue~\cite{huang2023take}. 
The work in~\cite{coffrin2014visualizing} analyzes student activity and success patterns. 
The study in~\cite{glance2014attrition} examines the duration of the course and the number of activities. 
Other works investigats the role of course design~\cite{ferguson2015examining} and learner demographics~\cite{el2017understanding}. 
These studies provide a foundation for understanding dropout behaviors comprehensively.

Predicting dropouts in MOOCs is extensively studied.
Some models focus on predicting dropout risks even before enrollment~\cite{li2023predicting}. Others leverage deep neural networks to predict dropout probabilities~\cite{imran2019predicting}. Additionally, supervised learning has been applied to detect potential dropouts in the early stages of learning~\cite{panagiotakopoulos2021early}. 

Intervention strategies aim to re-engage learners and reduce dropout rates in MOOCs. 
Personalization is a key approach~\cite{assami2018personalization}. 
Additionally, fostering social learning and interaction is shown to increase completion rates~\cite{crane2021influence}. 
Providing motivational support also plays a crucial role in recalling students~\cite{hew2014students}. 

\subsection{Dropouts in LLM-driven Interactive Online Courses}
LLMs have emerged as powerful tools in educational applications, reshaping the landscape of online and interactive learning~\cite{alsafari2024towards,rashid2024humanizing,bui2024cross,tan2025elf,lan2024teachers}.  
Recent advancements have demonstrated the potential of AI lecturers to simulate realistic classroom interactions, foster active engagement, and drive innovation in teaching practices~\cite{pang2024artificial,salminen2024using,zhang2024simulating}.  

Building upon recent advancements, LLM-driven interactive online courses, represented by Massive AI-empowered Courses (MAIC) redefine traditional MOOCs by creating interactive, text-based learning environments facilitated by LLM-driven multi-agent systems~\cite{yu2024moocmaicreshapingonline}.  
MAIC organizes learning around slides and lecture notes~\cite{zhang2024awaking}, with each slide serving as a unit of interaction, illustrated in Figure~\ref{fig:course-example}.  
Learners engage in dynamic conversations with various agents, including an AI Teacher, AI Teaching Assistants (TAs), and simulated peers, creating a collaborative and immersive in-class experience~\cite{zhang2024simulating}. 

Similar to traditional online courses, dropout issues in these new types of online courses also deserve attention and investigation.

%% file: paragraphs/3_dropouts_analysis.tex
\begin{figure*}[h]
    \centering
    \includegraphics[width=\linewidth]{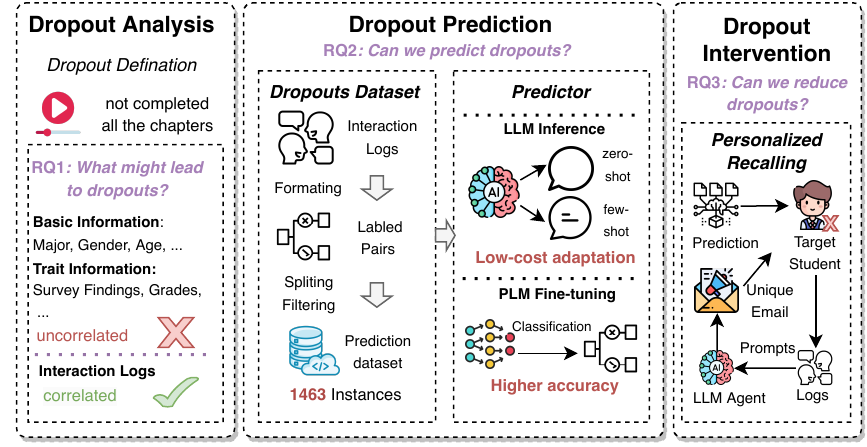}
    \caption{Workflow of dropout analyzing, predicting, and interventing in MAIC.}
    \label{fig:enter-label}
\end{figure*}

\section{MAIC Dropout Analysis}
To investigate dropouts in MAIC, we start with the data from a specific course, Towards Artificial General Intelligence (TAGI).
We define the dropouts in MAIC according to this data and analysis the relationships between the collected student's data and their dropout results.

\subsection{Data Source}
TAGI is a highly acclaimed in-person course offered at a top university in China.
Following the proposal of the MAIC model, with the collaboration of the course instructor, TAGI has been redesigned into an MAIC course, MAIC-TAGI.
MAIC-TAGI is divided into six major chapters to cover the history, present, and future directions of AI technology.
Currently, MAIC-TAGI has been offered for two semesters, with 186 students from various majors and grades participating in the course in the first semester, and ongoing with different students in the second semester.
These students' basic and trait information, and interaction logs at MAIC course support our subsequent research on dropouts.

\subsection{Correlation Analysis}
\label{sec:dropout-analysis}
\hspace{2em} \textcolor{maic-color}{\textbf{RQ~1:}} \textit{What might lead to dropouts?}

\begin{definition}[Chapter Study Complete]
Let \( C \) be the set of all chapters in a MAIC course. A chapter \( c \in C \) is considered ``Study Complete'' if and only if the teacher Agent has presented all the slides and lecture notes for \( c \). We denote the set of chapters that a student has completed as \( S \subseteq C \).
\end{definition}

\begin{definition}[Course Completion Progress]
Let \( |C| \) be the total number of chapters in a MAIC course. The Course Completion Progress \( P \) of a student is defined as the cardinality of the set \( S \), i.e., \( P = |S| \).
\end{definition}

\begin{definition}[Dropouts in MAIC]
A student is considered to have dropped out if \( S \neq C \), i.e., not all chapters are marked as complete.
\end{definition}

We collect and analyze the Course Completion Progress in MAIC-TAGI, as shown in Table~\ref{tb:completion}. 
We analyze the correlation between the Course Completion Progress and the students' information.
This information includes three categories: basic information, trait information, and interaction logs.

\begin{figure}[h]
    \centering
    \includegraphics[width=\linewidth]{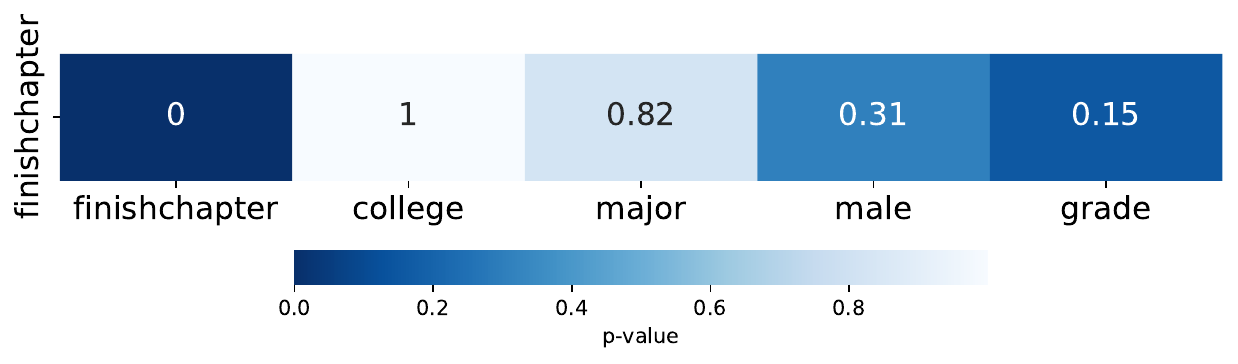}
    \caption{The Chi-square results between students' basic information and Course Completion Progress, indicating no significant correlation.}
    \label{fig:student-basic-information}
\end{figure}

\input{tables/completion}
\vpara{Basic Information: }
Figure~\ref{fig:student-basic-information} visualizes the correlation between students' college, major, gender, grade, and Course Completion Progress. 
A chi-square test examines the relationship among these qualitative variables. 
The results indicate no significant correlation.

\begin{figure}[h]
    \centering
    \includegraphics[width=\linewidth]{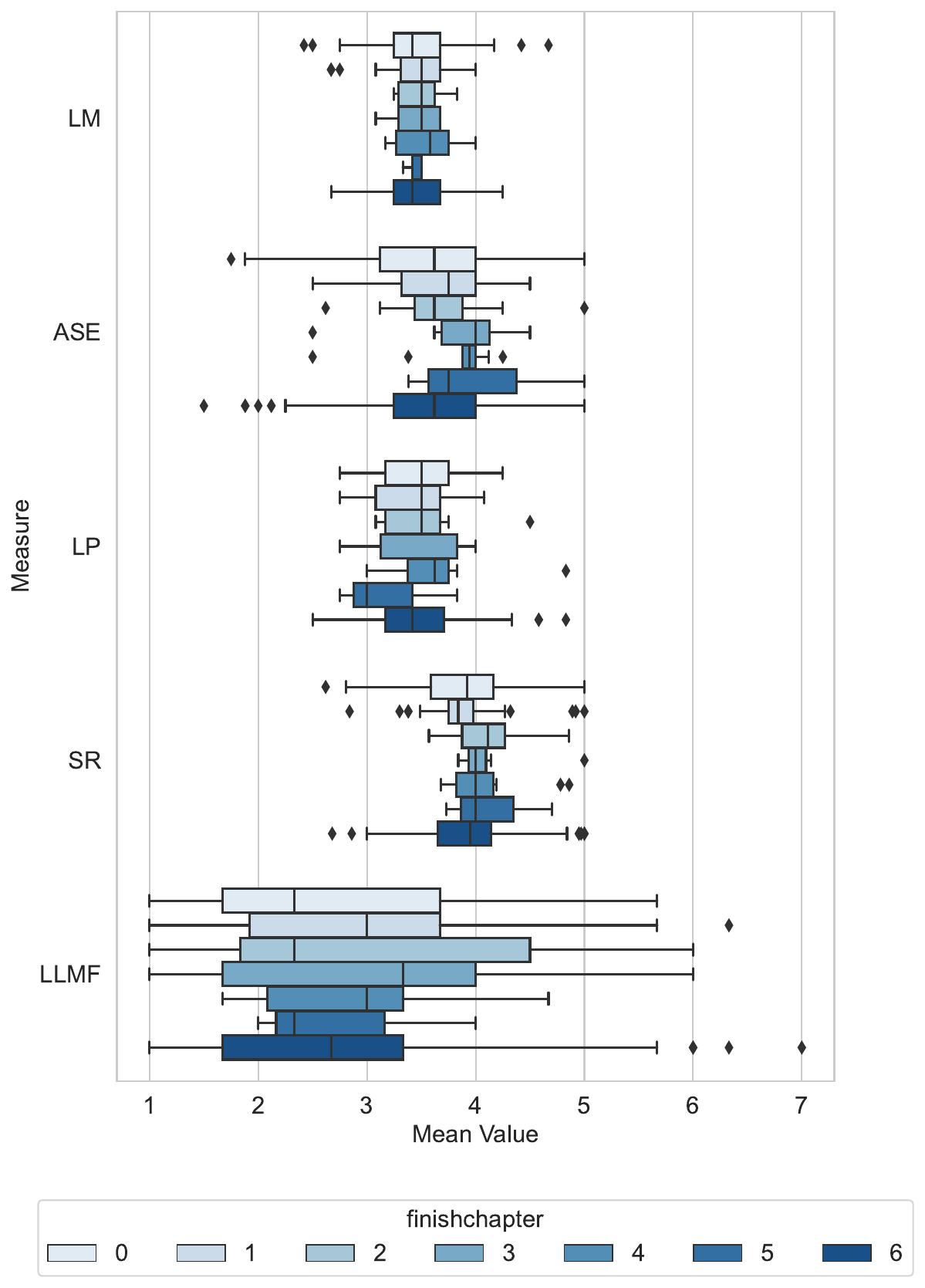}
    \caption{The Pearson results between students' trait information and Course Completion Progress, indicating low correlation.}
    \label{fig:student-trait-information}
\end{figure}

\vpara{Trait Information: }
At the start of each student's enrollment in the MAIC-TAGI course, they completed a questionnaire designed based on educational theories to assess their traits. 
This questionnaire measures five indicators: Learning Motivation (LM), Academic Self-Efficacy (ASE), Persistence (LP), Strategy (SR), and Large Model Usage Frequency (LLMF). 
Students rate themselves on a scale of 1 to 5 for each indicator. 
A Pearson correlation analysis examines the relationship between these indicators and Course Completion Progress, with the results visualized in Figure~\ref{fig:student-trait-information}. 
The analysis shows that the correlation between these traits and Course Completion Progress is low.

\begin{figure}[t]
    \centering
    \includegraphics[width=\linewidth]{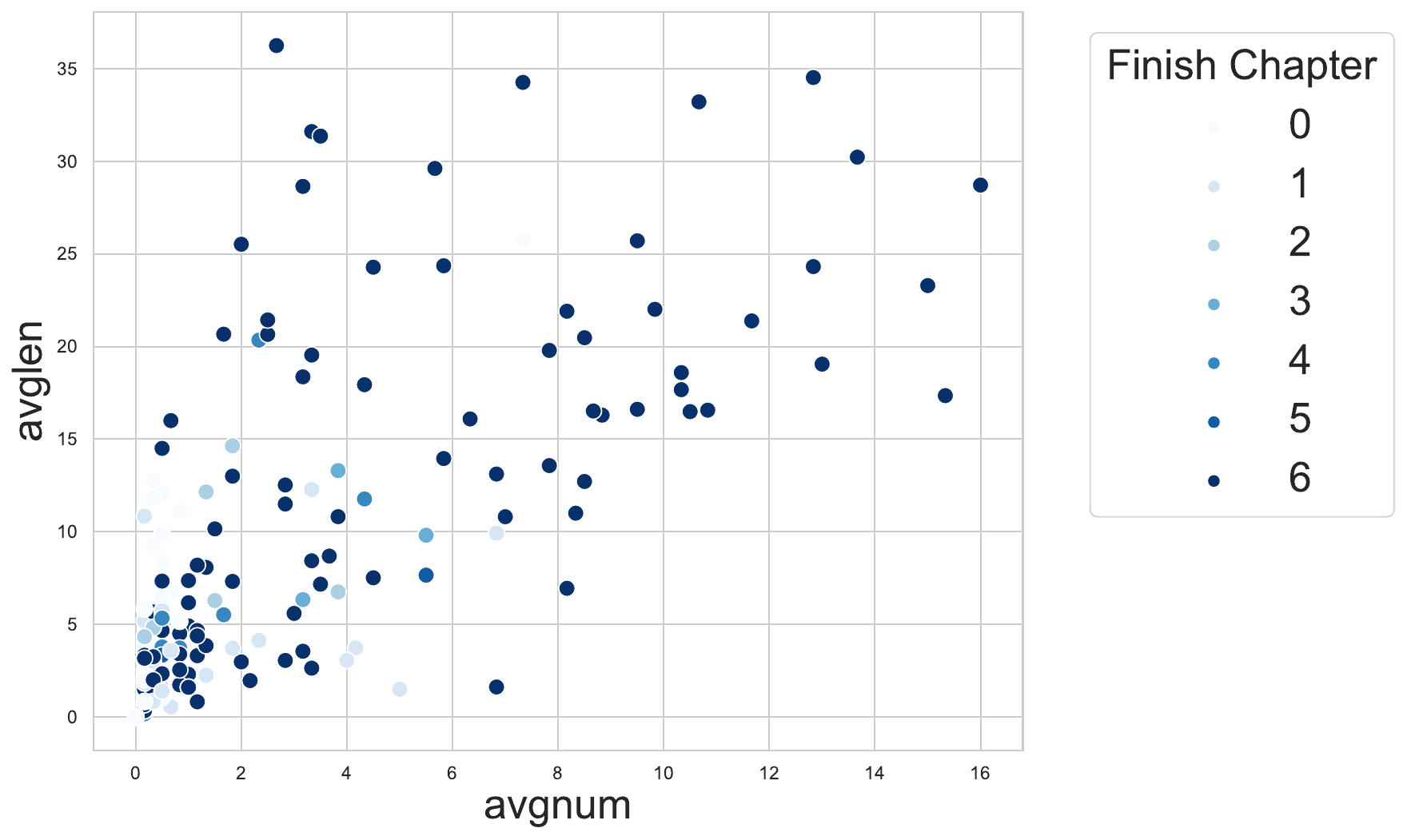}
    \caption{Visualization of the characteristics of the students' interaction information and Course Completion Progress.}
    \label{fig:interaction-information}
\end{figure}

\vpara{Interaction Logs: }
MAIC-TAGI collects all interaction records between students and the LLM-Agent in the classroom. 
Figure~\ref{fig:interaction-information} visualizes the characteristics of students with different Course Completion Progress based on the average number of interactions per chapter and the average length of each interaction.
In this coordinate system, points in the upper-right corner represent students who interact more actively with the system (both in frequency and length of interactions), while points closer to the origin indicate less proactive engagement (fewer interactions and shorter lengths). 
A clear trend emerges: students with higher completion rates (darker points) tend to exhibit more frequent and longer interactions, as reflected by their greater concentration in the upper-right corner compared to lighter points.

In summary, among the information currently collected by MAIC-TAGI, the most relevant factor to Course Completion Progress is the interaction records of students.

%% file: tables/completion.tex
\begin{table}[t!]
\centering
\small
\resizebox{\linewidth}{!}{
\begin{tabular}{c|ccccccc}
\toprule
\textbf{Chapter} & 0 & 1 & 2 & 3 & 4 & 5 & 6 \\
\midrule
\textbf{Students} & 34 & 22 & 8 & 3 & 7 & 2 & 110 \\
\bottomrule
\end{tabular}
}
\caption{Number of students completing each chapter in MAIC-TAGI. 76 students dropped out~(40.9\%).}
\label{tb:completion}
\end{table}

%% file: paragraphs/4_dropous_prediction.tex
\section{MAIC Dropout Prediction}
\label{sec:dropout_prediction}
\hspace{3em} \textcolor{maic-color}{\textbf{RQ~2:}} \textit{Can we predict dropouts?}

In MOOCs, dropout prediction aims to determine whether a student is likely to drop out in the near future based on their historical activities, including engagement with videos, forums, assignments, and webpage interactions. 
Similarly, this section defines the dropout prediction task in MAIC and proposes feasible solutions to address it.

\begin{figure}[t!]
    \centering
    \includegraphics[width=\linewidth]{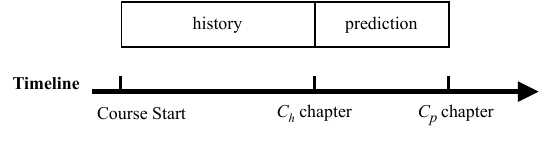}
    \caption{Timeline dropout prediction in MAIC.}
    \label{fig:dp-task}
\end{figure}

\subsection{Task Formulation}
The correlation analysis in Section~\ref{sec:dropout-analysis} suggests that interaction records are potential factors related to the likelihood of dropout. 
These interaction records comprise a series of structured strings documenting the textual content of interactions between students and multiple agents in the classroom, arranged chronologically. 
Accordingly, dropout prediction in MAIC can be defined as a binary classification task based on these structured strings.


\begin{definition}[Dropout Prediction in MAIC]
Given a student's interaction records with MAIC before the start of Chapter $C_h$ (during the history period), the task is to determine whether the student will drop out from the start of Chapter $C_h$ to the end of Chapter $C_p$ (during the prediction period):
\label{eq:dropout-prediction}
\begin{align}
\mathbb{P}(&\text{Dropout} \mid \mathcal{I}_{C_h}, C_h, C_p), \nonumber \\ 
\text{where} \quad &C_h \in [1, \text{Last Chapter}], \nonumber \\ 
&C_p \in [C_h, \text{Last Chapter}]. \nonumber
\end{align}
\end{definition}
\noindent Here, $\mathcal{I}_{C_h}$ denotes the interaction records before Chapter $C_h$, $C_h$ represents the start of the history period, and $C_p$ represents the end of the prediction period, as illustrated in Figure~\ref{fig:dp-task}.
A more detailed explanation of the valid combinations of \(C_h\) and \(C_p\) is provided in the Appendix~\ref{apdx:valid-combination}.

\subsection{Dropout Dataset}
The interaction records in the MAIC-TAGI course are organized into a dataset by varying the values of \( C_h \) and \( C_p \). 
The input consists of \( C_h \), \( C_p \), and the interaction records preceding \( C_h \), while the output is a binary label indicating whether a course withdrawal occurred.

In MAIC-TAGI, \( C_h \) ranges from 1 to 6, and \( C_p \) ranges from \( C_h \) to 6. A student's class record can generate multiple dropout prediction instances by combining different values of \( C_h \) and \( C_p \). Under these constraints, we extracted 1201 dropout prediction instances from the class records of 186 students. Of these, 20\% are used as the test set, and the remainder as the training set.

Detailed examples of valid combinations for dropout prediction and the full dataset statistics are provided in Appendix~\ref{appendix:dropouts-dataset}.

\subsection{Adaptive Predictor}
The dropout prediction task in MAIC is a binary classification task for structured text. To handle different stages of course progress and varying amounts of interaction records, we propose a Course-progress-adaptive Dropout Prediction Framework (CPADP). This framework employs different prediction methods at different stages:

\textbf{1. Zero-Shot Stage:} At the beginning of the course, zero-shot methods with large pre-trained language models (LLMs) assess dropout risk using prior knowledge and basic understanding of interaction data.

\textbf{2. Few-Shot Stage:} As records accumulate, labeled examples are incorporated, and prompt engineering is applied to enhance LLM performance.

\textbf{3. Fine-Tuning Stage:} With sufficient data, smaller pre-trained language models (PLMs) are fine-tuned for binary classification. Features from interaction records \(\mathcal{I}_{C_h}\) are extracted using a PLM, and a Multi-Layer Perceptron (MLP) predicts dropout probabilities.
This adaptive framework ensures accurate predictions while maintaining computational efficiency across multiple courses.

\subsection{Prediction Performance}
We evaluate prediction performance using F1 Score and Accuracy. 
Table~\ref{tb:predictors} summarizes results averaged over 3 tests with different random seeds. 
GPT-4 with Few-shot inference achieves the best performance among LLMs, while PLM fine-tuning surpasses all LLM methods. 
The best-performing prompt is shown in Figure~\ref{fig:prompt}.
Appendix~\ref{apdx:different-prompt-designs} provides a comparison of the inference results among different prompt designs.
\input{tables/predictor_result}

While the PLM method performs well, it requires substantial training data. 
For new courses with limited interaction records and unlabeled data, predictions rely on zero-shot or few-shot settings. 
CPADP dynamically adapts as the course progresses, improving over time. Once sufficient data is available, PLMs can be consistently used, including across semesters of the same course.

%% file: tables/predictor_result.tex
\begin{table}[t!]
\centering
\small
\resizebox{\linewidth}{!}{
\begin{tabular}{c|c|cccc}
\toprule
Method & Setting & Precision & Recall & F1\_score & Accuracy \\
\midrule
\multirow{2}{*}{GPT-4} & ZS & 0.909 & 0.250 & 0.392 & 0.716 \\
                       & FS & 0.921 & 0.450 & 0.604 & 0.779 \\
\midrule
\multirow{2}{*}{GLM-4} & ZS & 0.897 & 0.231 & 0.366 & 0.704 \\
                       & FS & 0.812 & 0.400 & 0.536 & 0.742 \\
\midrule
\multirow{2}{*}{DeepSeek} & ZS & 0.897 & 0.231 & 0.366 & 0.704 \\
                          & FS & 0.562 & 0.430 & 0.487 & 0.662 \\
\midrule
PLM & FT & 0.966 & 0.906 & 0.935 & 0.954 \\
\bottomrule
\end{tabular}
}
\caption{Dropout predictors result. ZS indicates Zero Shot, FS indicates Few Shots, FT indicates Fine Tuning.}
\label{tb:predictors}
\end{table}

%% file: paragraphs/5_dropous_intervention.tex
\section{MAIC Dropout Intervention}
\hspace{3em} \textcolor{maic-color}{\textbf{RQ~3:}} \textit{Can we reduce dropouts?}

The prediction of dropouts aims to identify at-risk dropping-out students.
For these students, drawing inspiration from products like Duolingo~\footnote{\url{https://en.wikipedia.org/wiki/Duolingo}} and Character.AI~\footnote{\url{https://en.wikipedia.org/wiki/Character.ai}}, which leverage personalized notifications to re-engage users, we utilized these students' interaction records to generate personalized reminder Emails using LLM, helping them recall engaging class content and reignite their interest in returning. 
This approach establishes a complete ``prediction-recognition-intervention" loop based on student interaction data in MAIC.

\subsection{Generating Personalized Emails}
By creating an Email agent using the prompt in Figure~\ref{fig:prompt}, we generate engaging, personalized content that not only recalls past learning moments but also uses humor, inspiration, or a conversational tone to re-engage students. 

\begin{figure}[t]
    \centering
    \includegraphics[width=\linewidth]{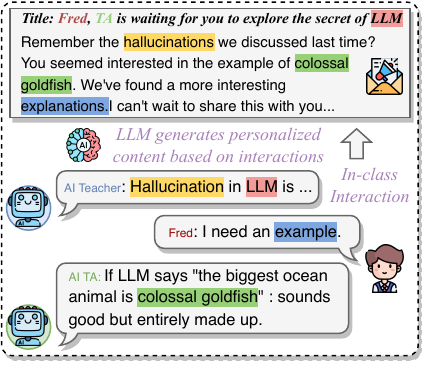}
    \caption{An example of personalized Email generation.}
    \label{fig:email}
\end{figure}

For example, as shown in Figure~\ref{fig:email}, if a student named \textit{``Fred"} was confused about the concept of \textit{``Hallucination"}, the Email might reference the earlier session on \textit{``General Artificial Intelligence Overview"} and preview upcoming contents according to the interaction logs and the topic which Fred may be interested in. 

\subsection{Intervention Results}
On January 10, the 65th day in the middle of the new semester in the MAIC-TAGI course, we predict students' dropout probabilities using a setting where both $C_h$ and $C_p$ are the chapters they are currently studying.
Based on the prediction results, we generate and send personalized emails to the at-risk students. 

We compare the number of logins before and after the intervention: 14 logins during Days 63--65 versus 25 logins during Days 66--68, demonstrating a clear increase in student login frequency and engagement as a direct result of our intervention.

To confirm that the increase is primarily driven by our intervention, we analyze the 17 students who log in during Days 66--68. 
Specifically, we assess whether the intervention prompts students who would otherwise not log in to engage with the course, ruling out other factors such as increased availability over the weekend. 
These students are divided into two groups: 9 self-initiated (non-intervened) and 8 recalled (intervened).

\input{tables/intervention_results}

Table~\ref{tab:intervention_analysis} shows that the recalled group has significantly higher offline days (52.6 days on average) and minimal engagement before the intervention, with chapter progress of 0.75 and very few posts (both in number and length). 
In contrast, the self-initiated group has much fewer offline days (7.66 days on average), higher chapter progress (2.56), and more active participation.
These results indicate that our predictions and interventions precisely recalled students who were not actively engaged in interactions, contributing to the observed increase in logins and learning activity.

%% file: tables/intervention_results.tex
\begin{table}[t!]
\centering
\footnotesize 
\begin{tabularx}{\linewidth}{@{}lXXXX@{}} 
\toprule
\textbf{Group (headcount)} & \textbf{Offline Days} & \textbf{Chapter} & \textbf{Msg Num} & \textbf{Msg Length} \\ \midrule
Self-initiated (9)  & 7.66 & 2.56 & 8.78 & 106.56 \\
Recalled (8)        & 52.6 & 0.75 & 0.25 & 0.75 \\ \bottomrule
\end{tabularx}
\caption{Comparison of students' activities by intervention status. Chapter represents the average course chapter progress, Msg Num indicates the average number of messages sent, and Msg Length refers to the average number of words per message.}
\label{tab:intervention_analysis}
\end{table}

%% file: paragraphs/6_conclusion.tex
\section{Conclusion}
\label{sec:conclusion}
This paper studies dropouts in Massive AI-empowered Courses (MAIC) through an empirical examination on the MAIC-TAGI course.
We propose the CPADP framework and a personalized email system to predict and recall at-risk students.
Our discovery highlights the effectiveness of combining interaction data, adaptive prediction models, and personalized interventions to handle dropouts in LLM-driven online learning environments. 

%% file: paragraphs/7_limitations_and_ecs.tex
\section{Limitations}

As this paper represents an initial exploration into dropout analysis in a novel AI-driven online course setting, several limitations remain in our methods and findings. 

\vpara{Course Diversity: }
The dropout analysis, prediction, and email recall validation in this work were conducted using data from a single MAIC course, \textit{Towards Artificial General Intelligence (TAGI)}, offered at a top university in China. 
Due to the complexity of data collection and analysis processes, the study lacks diversity in terms of the courses used for evaluation. Future work should expand the scope to include more MAIC courses across different subjects, difficulty levels, and institutions to ensure the generalizability of the proposed methods.

\vpara{Experimental Completeness: }
The evaluation of the personalized email recall system is relatively limited in scope. 
Future studies will design more comprehensive, diverse, and robust experiments. 
This could include A/B testing with larger student populations, varying email designs, and longer-term tracking of re-engagement effects.

\section{Ethical Considerations}
\vpara{Data Privacy and Informed Consent}  
All interaction data is anonymized to safeguard participant privacy and confidentiality. 
Participants are fully informed about the study’s purpose, the use of AI-generated content, and the data collection process. 
Informed consent is obtained, ensuring participants understand their rights and can withdraw at any time.

\vpara{Accuracy of AI-Generated Content}  
LLM-based educational systems, like the one in this study, may occasionally generate incorrect or misleading information. Despite carefully designed content and agent responses, these risks cannot be fully eliminated. To mitigate potential issues, we actively monitor AI outputs and provide corrections or clarifications to students when inaccuracies arise.

\vpara{Ethical Deployment of LLM Systems}  
The deployment of LLM-powered systems in education requires thoughtful consideration of their impacts. This study is conducted in a controlled academic setting, and its findings are not applied to broader real-world contexts without further validation. Future work will emphasize additional safeguards and thorough evaluations to ensure these systems are both ethical and effective in diverse scenarios.

%% file: paragraphs/8_appendix.tex
\newpage
\begin{figure*}[t!]
    \centering
    \includegraphics[width=\linewidth]{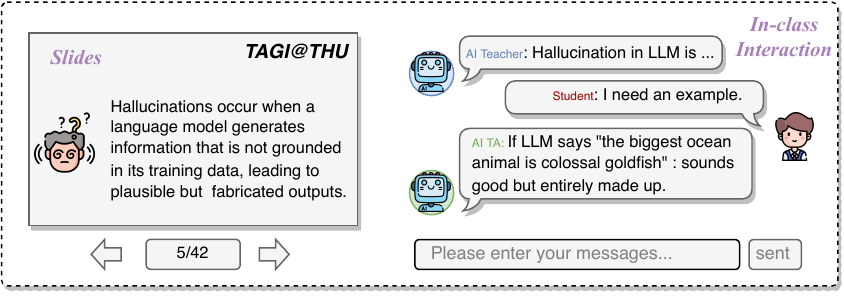}
    \caption{A schematic diagram of the learning environment in MAIC.}
    \label{fig:course-example}
\end{figure*}
\appendix
\section{The Deployed MAIC System}
As mentioned in the paper, the MAIC system has been deployed for over one year, supporting dozens of courses, across more than 3,000 students. 
Due to the double-blind review requirement, we are unable to provide further details about the deployed system at this time.

\section{Learning Modes in the MAIC Course}
In the MAIC classroom, students engage in a highly personalized and interactive learning environment, as depicted in Figure~\ref{fig:course-example}. 
Students interact with different types of AI agents: AI teachers, AI teaching assistants, and other customizable AI classmates. 
These agents collaboratively create a dynamic and personalized learning experience that adapts to individual student needs.

\section{Case Study of Interaction Logs}
\begin{figure*}[h]
    \centering
    \includegraphics[width=\linewidth]{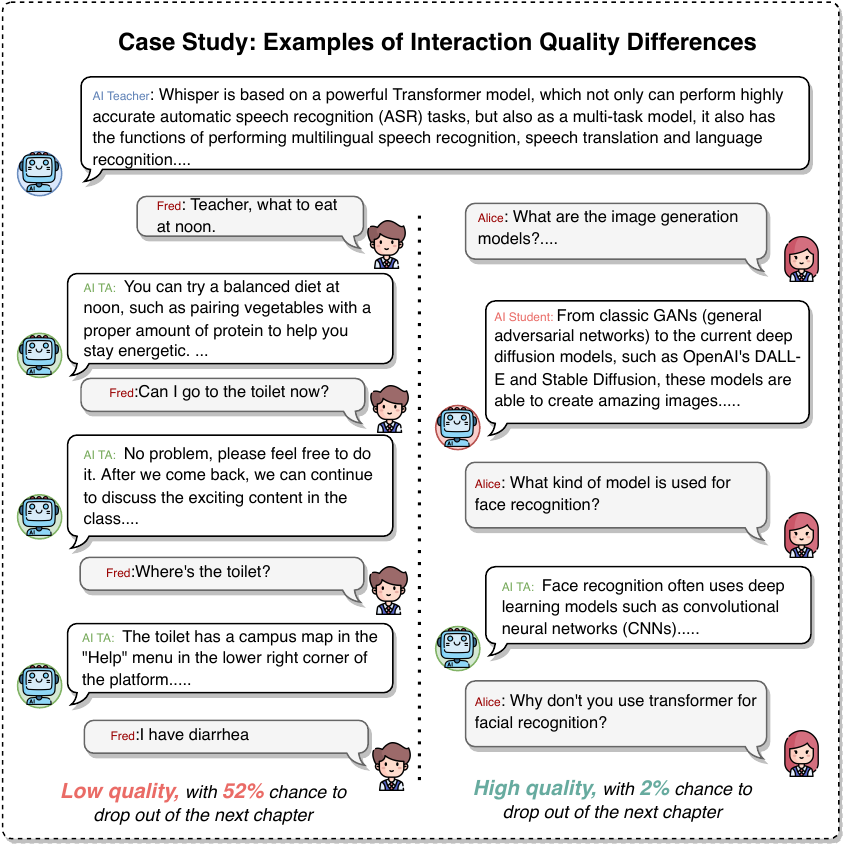}
    \caption{Case Study~(A pseudonym is used, but the case is real)}
    \label{case}
\end{figure*}

In the MAIC classroom, students' interaction logs are closely related to their class participation, and the probability of dropping out. 
To illustrate this relationship, a case is constructed, as shown in Figure \ref{case}. 
In this case, Fred's irrelevant interactions resulted in a high dropout probability, with a predicted likelihood of 52\% for the subsequent chapter. 
In contrast, Alice's highly relevant interactions, indicating active engagement in the class, yielded a much lower dropout probability of only 2\%. 
Notably, the actual outcomes aligned with these predictions: Fred dropped out in the next chapter, while Alice remained enrolled.

\section{Details of the Dropout Dataset}
\label{appendix:dropouts-dataset}
In MAIC-TAGI, the dropout dataset is constructed by varying the values of \( C_h \) and \( C_p \) for each student's interaction records. The input consists of \( C_h \), \( C_p \), and the interaction records preceding \( C_h \), while the output is a binary label indicating whether a course withdrawal occurred.
\subsection{Dataset Statistics}
Table~\ref{tb:dataset} provides detailed statistics of the dropout dataset. A total of 1201 dropout prediction instances were generated from the class records of 198 students, with 20\% used as the test set and the remaining 80\% as the training set.

\input{tables/dataset}
\input{tables/case}

\subsection{Valid Combinations of \( C_h \) and \( C_p \)}
\label{apdx:valid-combination}
It is noteworthy that $C_h$ cannot be zero because prediction is meaningless when there are no records. 
The value of $C_p$ can also equal $C_h$, meaning that all historical interaction data before the start of this chapter is used to predict whether the student will drop out by the end of the same chapter.

For example, if a student drops out during Chapter 3, the valid combinations of \( C_h \) and \( C_p \) are:
{\small
\[
(1, 1), (1, 3), (2, 2), (2, 3), (3, 3), (4, 4), (5, 5), (6, 6).
\]
}
Table~\ref{tb:case} shows the instances generated from this case.
The $\checkmark$ marks the label as True, which means this student drops out.

\subsection{Case-aware Prediction Accuracy}
\begin{figure}[h]
    \centering
    \includegraphics[width=\linewidth]{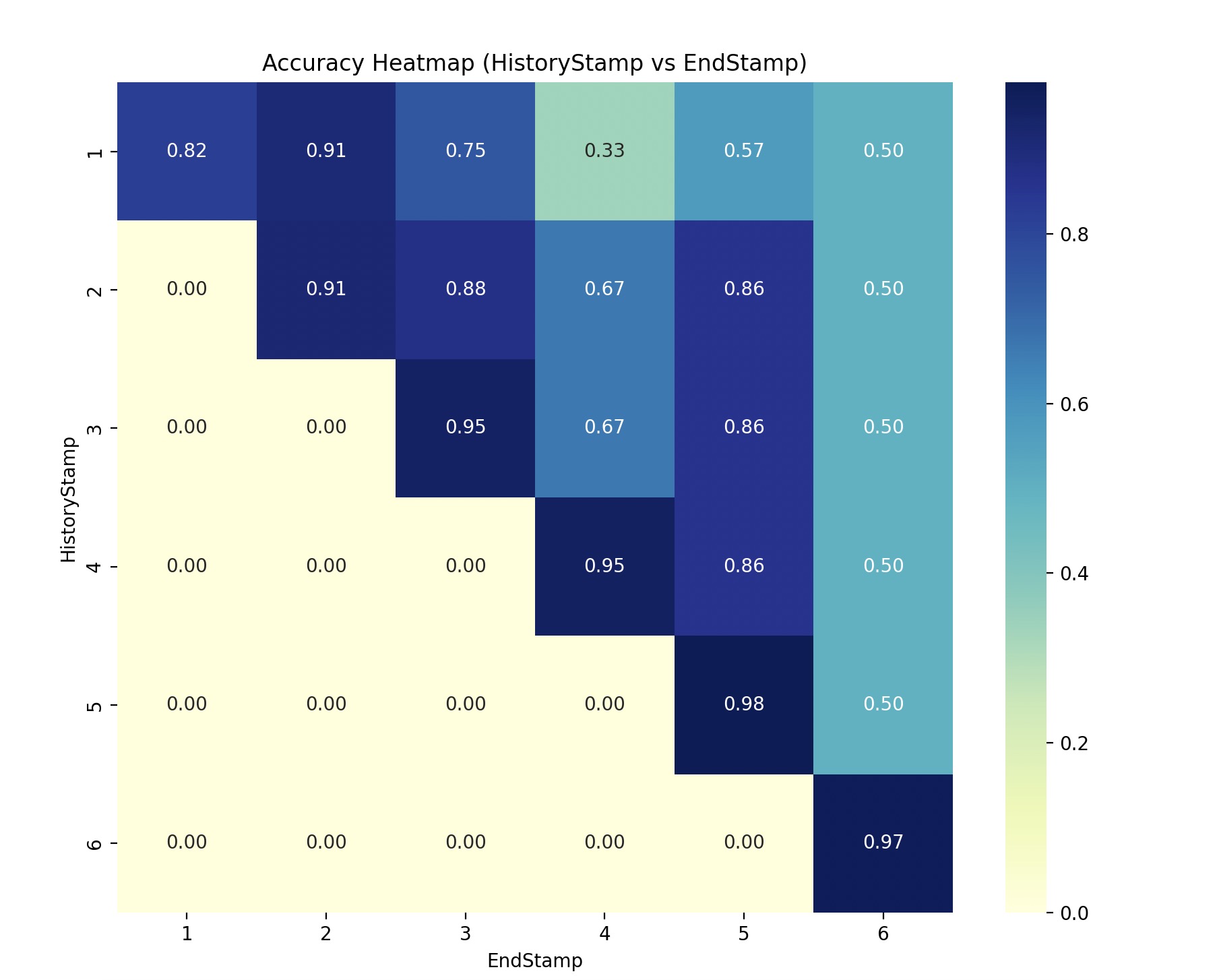}
    \caption{Case accuracy details heatmap. ``HistoryStamp'' corresponds to $C_h$, and ``EndStamp'' corresponds to $C_p$. }
    \label{fig:case-acc}
\end{figure}

Section~\ref{sec:dropout_prediction} presents the overall prediction accuracy of our proposed CPADP framework. Figure~\ref{fig:case-acc} provides a detailed visualization of the case-level accuracy across the entire dataset. 

It can be observed that the prediction accuracy decreases as the difference $C_p - C_h$ increases from the results. 
This trend suggests that the larger the gap between the student's current progress and their historical record, the more challenging it becomes to make accurate predictions.

Therefore, when applying dropout prediction in practical classroom settings, the optimal configuration is to base predictions on the student’s current chapter ($C_p$) and their performance in previous chapters ($C_h$). 
This approach ensures a more accurate assessment of whether the student is likely to complete the current chapter.

\begin{figure*}[h]
    \centering
    \includegraphics[width=\linewidth]{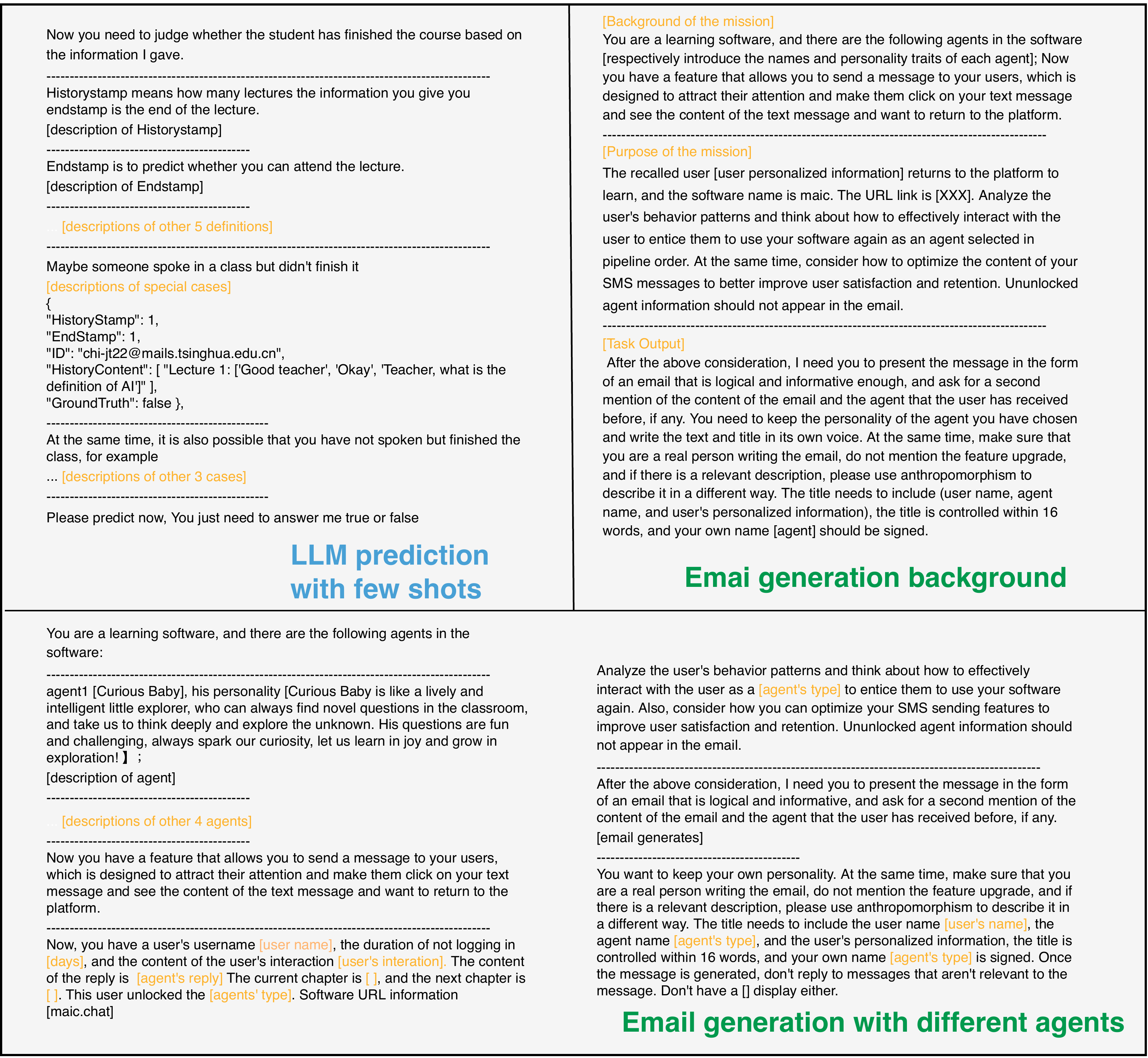}
    \caption{Prompts used in prediction and Email generation.}
    \label{fig:prompt}
\end{figure*}


\section{Details of Dropout Prediction}
\subsection{Prompt Designs}
\label{apdx:different-prompt-designs}
Table~\ref{tb:few-shot} lists the accuracy of GPT-4 under zero-shot settings and four different few-shot settings. 
The results show that including diverse examples, such as special cases and causal examples, significantly improves performance.
\input{tables/few-shot}

\subsection{PLM Structure}
To predict dropout probabilities, the interaction records \(\mathcal{I}_{C_h}\) are first processed through a pre-trained language model (PLM) to extract features. The resulting features are then passed into a Multi-Layer Perceptron (MLP) for binary classification. The process can be formally described as follows:
\begin{align}
\mathbf{h} &= \text{PLM}(\mathcal{I}_{C_h}), \nonumber \\
\mathbb{P}(\text{Dropout}), &\mathbb{P}(\text{Retention}) = \text{MLP}(\mathbf{h}). \nonumber
\end{align}
Here, \(\mathbf{h}\) represents the feature vector generated by the PLM, while \(\mathbb{P}(\text{Dropout})\) and \(\mathbb{P}(\text{Retention})\) are the predicted probabilities of dropout and retention, respectively.

%% file: tables/dataset.tex
\begin{table}[t!]
\centering
\small
\resizebox{\linewidth}{!}{
\begin{tabular}{cc|cccccc}
\toprule
\multicolumn{2}{c|}{} & \multicolumn{6}{c}{$C_p$} \\
& & 1 & 2 & 3 & 4 & 5 & 6 \\
\midrule
\multirow{6}{*}{$C_h$}& 1 & 186 & 22 & 8 & 3 & 7 & 2 \\
&2 & 0 & 186 & 8 & 3 & 7 & 2 \\
&3 & 0 & 0 & 186 & 3 & 7 & 2 \\
&4 & 0 & 0 & 0 & 186 & 7 & 2 \\
&5 & 0 & 0 & 0 & 0 & 186 & 2 \\
&6 & 0 & 0 & 0 & 0 & 0 & 186 \\
\bottomrule
\end{tabular}
}
\caption{Scale of MAIC-TAGI dropouts prediction dataset.}
\label{tb:dataset}
\end{table}

%% file: tables/case.tex
\begin{table}[t!]
\centering
\small
\resizebox{\linewidth}{!}{
\begin{tabular}{cc|cccccc}
\toprule
\multicolumn{2}{c|}{} & \multicolumn{6}{c}{$C_p$} \\
& & 1 & 2 & 3 & 4 & 5 & 6 \\
\midrule
\multirow{6}{*}{$C_h$}& 1 & $\times$ &  & $\checkmark$ &  &  &  \\
&2 &  & $\times$ & $\checkmark$ &  &  &  \\
&3 &  &  & $\checkmark$ &  &  &  \\
&4 &  &  &  & $\checkmark$ &  &  \\
&5 &  &  &  &  & $\checkmark$ &  \\
&6 &  &  &  &  &  & $\checkmark$ \\
\bottomrule
\end{tabular}
}
\caption{Dropouts instances from a specific student who drops out at Chapter 3. The $\checkmark$ marks the label as True, while the $\times$ marks the label as False.}
\label{tb:case}
\end{table}

%% file: tables/few-shot.tex
\begin{table}[t!]
\centering
\small
\resizebox{\linewidth}{!}{
\begin{tabular}{c|c}
\toprule
Prompt Design & Accuracy \\
\midrule
Zero-shot & 0.716 \\
Random Selecting Examples & 0.730 \\
Only False Examples & 0.745 \\
Two Special Case Examples & 0.760 \\
Special Case Examples and Casual Examples & 0.779 \\
\bottomrule
\end{tabular}
}
\caption{Dropout predictors accuracy of GPT-4 under different prompt settings.}
\label{tb:few-shot}
\end{table}